\documentclass[runningheads]{llncs}
\usepackage[T1]{fontenc}
\usepackage{graphicx,verbatim}

\usepackage{amsmath}
\usepackage{amssymb}
\usepackage{booktabs}
\usepackage{multirow}
\usepackage{soul}
\usepackage{dsfont}
\usepackage[table,dvipsnames]{xcolor}
\usepackage{bm}

\usepackage[T1]{fontenc}
\usepackage{graphicx}
\usepackage{color}
\usepackage{mathtools}
\usepackage{amsfonts}
\usepackage[labelfont=bf]{caption}
\usepackage[labelformat=simple,labelfont=bf]{subcaption}

\usepackage{booktabs,multirow}
\usepackage{array}
\usepackage{pifont}
\usepackage{bbding}
\usepackage[pagebackref=true,breaklinks=true,colorlinks=true,bookmarks=false,citecolor=blue,urlcolor=blue,linkcolor=blue]{hyperref}
\usepackage[misc]{ifsym}

\definecolor{yellow}{rgb}{1,0.97, 0.65}
\definecolor{lightyellow}{rgb}{1,1, 0.8}
\definecolor{orange}{rgb}{1, 0.85, 0.7}
\definecolor{tablered}{rgb}{1, 0.7, 0.7}

\usepackage[capitalize]{cleveref}
\crefname{section}{Sec.}{Secs.}
\Crefname{section}{Section}{Sections}
\Crefname{table}{Table}{Tables}
\crefname{table}{Tab.}{Tabs.}

\newcommand{\boldstartspace}[1]{\vspace{0.1in}\noindent\textbf{#1}}

\usepackage{xargs}
\usepackage[colorinlistoftodos,prependcaption,textsize=tiny]{todonotes}
\newcommandx{\Shi}[2][1=]{\todo[linecolor=red,backgroundcolor=red!25,bordercolor=red,#1]{#2}}
\newcommandx{\change}[2][1=]{\todo[linecolor=blue,backgroundcolor=blue!25,bordercolor=blue,#1]{#2}}
\newcommandx{\info}[2][1=]{\todo[linecolor=OliveGreen,backgroundcolor=OliveGreen!25,bordercolor=OliveGreen,#1]{#2}}
\newcommandx{\improvement}[2][1=]{\todo[linecolor=Plum,backgroundcolor=Plum!25,bordercolor=Plum,#1]{#2}}
\newcommandx{\thiswillnotshow}[2][1=]{\todo[disable,#1]{#2}}

\begin{document}

\title{ClipGS: Clippable Gaussian Splatting \\for Interactive Cinematic Visualization of \\Volumetric Medical Data}


\author{
    Chengkun Li\inst{1} \and
    Yuqi Tong\inst{1} \and
    Kai Chen\inst{1} \and Zhenya Yang\inst{1} \and Ruiyang Li\inst{1} \and\\
    Shi Qiu\inst{1} \and Jason Ying-Kuen Chan\inst{2} \and Pheng-Ann Heng\inst{1} \and
    Qi Dou\inst{1}\textsuperscript{(\Letter)}
}
\authorrunning{C. Li et al.}
%
\institute{
    Department of CSE, The Chinese University of Hong Kong \and
    Department of ENT, The Chinese University of Hong Kong
}


\maketitle

\begin{abstract}

The visualization of volumetric medical data is crucial for enhancing diagnostic accuracy and improving surgical planning and education. Cinematic rendering techniques significantly enrich this process by providing high-quality visualizations that convey intricate anatomical details, thereby facilitating better understanding and decision-making in medical contexts. However, the high computing cost and low rendering speed limit the requirement of interactive visualization in practical applications.
In this paper, we introduce ClipGS, an innovative Gaussian splatting framework with the clipping plane supported, for interactive cinematic visualization of volumetric medical data.
To address the challenges posed by dynamic interactions, we propose a learnable truncation scheme that automatically adjusts the visibility of Gaussian primitives in response to the clipping plane. 
Besides, we also design an adaptive adjustment model to dynamically adjust the deformation of Gaussians and refine the rendering performance.
We validate our method on five volumetric medical data (including CT and anatomical slice data), and reach an average $36.635~\text{PSNR}$ rendering quality with $156~\text{FPS}$ and $16.1\text{MB}$ model size, outperforming state-of-the-art methods in rendering quality and efficiency. Project is available at: \href{https://med-air.github.io/ClipGS/}{https://med-air.github.io/ClipGS}.

\keywords{Volumetric Medical Data Visualization \and Interactive Cinematic Rendering \and Clippable Gaussian Splatting.}

\end{abstract}    
\section{Introduction}


Visualization of volumetric medical data~\cite{zhang2011volume} is a cornerstone of modern medical practice. The effectiveness of medical volume visualization is influenced by two key factors. First, the quality of rendering directly impacts the richness of the information conveyed. Recently, high-quality cinematic rendering technology~\cite{glemser2018new} is widely used in surgical planning~\cite{steffen2022three}, medical education~\cite{jabbireddy2023accelerated}. Second, the level of interactivity affects the user's understanding of the data. It enables the intuitive visualization of complex medical data while effectively highlighting anatomical abnormalities, facilitated by interactive tools such as clipping planes~\cite{binder2019leveraging,eid2017cinematic}.


To realize cinematic-level rendering of complex anatomical structures, ray tracing~\cite{lafortune1993bi} is widely to simulate the light propagation between voxels, enhancing detail and achieving high-quality photorealism~\cite{comaniciu2016shaping,veach1995bidirectional}. To extend the usability of cinematic volume medical data visualization in practical applications, researchers ~\cite{binder2019leveraging,eid2017cinematic,elshafei2019comparison} introduce clipping plane and clipping cube to interactively show the internal anatomy structures. However, the time-consuming rendering limits its practical application. 
To accelerate rendering processing, photon mapping-based volume rendering~\cite{yuan2024cinematic} is proposed in volumetric medical data visualization.  Despite its advantages, the computational demands of this method still restrict real-time performance on consumer-grade hardware. 
Although Heinrich~\cite{heinrich2020interacting} and Taibo~\cite{taibo2024immersive} further optimize the efficiency of interactive 3D medical visualization, their methods have to compromise rendering quality to achieve real-time interactions with volume clipping. 
It remains an open challenge to design a more efficient solution, implementing photo-realistic visualization while simultaneously supporting real-time interaction.

Recent advancements in neural rendering techniques have achieved great success in photo-realistic rendering with novel views. Neural radiance fields (NeRF)~\cite{mildenhall2021nerf} represent 3D scenes as a continuous volumetric function through a multi-layer perception (MLP). More recently, 3D Gaussian Splatting (3DGS) models the 3D scene as a set of anisotropic Gaussian primitives~\cite{kerbl3Dgaussians,10896112,10972671,zhu2025rethinking} to implement the point-based radiance rendering process, achieving high-quality rendering and achieves impressive real-time performance. Benefiting from these,  Niedermayr et al.~\cite{niedermayr2024application} explored the potential capacity of 3DGS~\cite{kerbl3Dgaussians} technology for anatomical visualization. Kleinbeck et al.~\cite{kleinbeck2024multi} took this a step further by extending a layered GS representation to visualize different anatomical structures incrementally. Although they implement real-time cinematic rendering of volumetric medical data, these methods only support rendering the fixed 3D structure and lack the ability to show internal structures and details. The non-interactive rendering limits the potential in surgical planning and education.

In this paper, we aim to establish a framework for real-time interactive cinematic visualization of volumetric medical data based on Gaussian Splatting. Here, we leverage the clipping plane, which is widely used in medical practice~\cite{binder2019leveraging,eid2017cinematic,elshafei2019comparison}, to interactively visualize the internal structures and highlight the details. 
However, the use of a clipping plane introduces an additional variable, making it difficult to directly adapt to 3DGS-based rendering.
Unlike existing method~\cite{kleinbeck2024multi} that constructs 3D Gaussians for each clipping layer and exponentially increases storage/memory cost, we propose a novel GS-based framework that can render volumetric medical data with photorealism and clipping plane interaction in real-time.
We introduce a learnable attribute to the Gaussian primitive to automatically control the visibility with the query clipping plane. Besides, we suggest dynamically adjusting the position and shape of visible Gaussian under the specific clipping plane to refine the rendering performance in the clipping surface.
Our main contributions are summarized as follows:
\begin{itemize}
    \item We propose ClipGS, a novel GS-based framework for real-time interactive cinematic anatomy visualization. 
    \item We propose a learnable truncation (LT) scheme, which can dynamically control the visibility of each Gaussian primitive with the query clipping plane. 
    \item We design an adaptive adjustment model (AAM) to adjust the deformation of Gaussian to refine the rendering performance.
    \item We build a cinematic medical dataset to validate our proposed method, demonstrating its superiority in visual quality and efficiency.
\end{itemize}
\begin{figure}[t]
    \centering
    \includegraphics[width=0.99\linewidth]{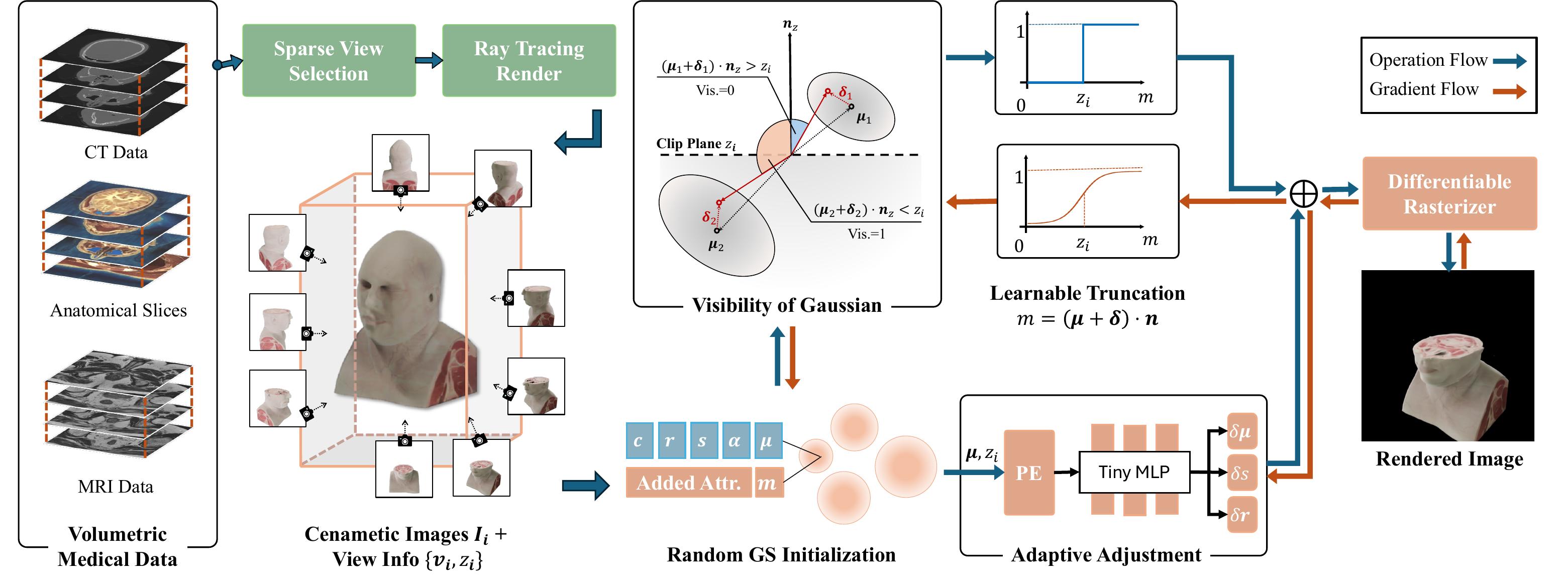}
    \caption{\textbf{Overview of our proposed framework.} Our ClipGS implements a real-time interactive cinematic rendering under the query condition (view direction \& clipping plane) for a given volumetric medical data.}
    \label{fig:pipline}
\end{figure}

\section{Method} \label{sec:method}

The overview pipeline of our proposed framework is shown in Fig.~\ref{fig:pipline}. For a given volumetric medical data $\mathbb{V}$, our method implements a real-time cinematic rendering for a given query view and clipping plane. Firstly, we use a ray-tracing renderer to generate a sparse cinematic image sequence $\{\bm{I}_i\}^N_{i=1}$ under random camera views $\bm{v}_i$ and random clipping plane $\bm{z}_i$. Then, we optimize a random initialized 3D Gaussian point cloud (Sec.~\ref{sec:3dgs}) from these images. For doing this, We introduce a learnable attribute to the Gaussian primitive to automatically control the visibility with the query clipping plane (Sec.~\ref{sec:truncation}). To refine the rendering quality, we design an adaptive adjustment model to dynamically adjust the position and shape of visible Gaussian under the specific clipping plane (Sec.~\ref{sec:refinement}). Our experiments show the superiority of the proposed method in rendering quality and efficiency (Sec.~\ref{sec:experiments}).

\subsection{Preliminaries of Gaussian Splatting} 
\label{sec:3dgs}
3D Gaussian splatting~\cite{kerbl3Dgaussians} represents the static 3D scene with a set of Gaussian primitives. Mathematically, it could be written as:
\begin{equation} \small
    \mathcal{G}(\bm{x};\bm{\mu},\bm{r},s) = e^{-\frac{1}{2} \left( \bm{x}-\bm{\mu} \right)^T \bm{\Sigma}^{-1} \left( \bm{x}-\bm{\mu} \right)}, \ \ \bm{\Sigma}= \bm{R}\bm{S}\bm{S}^T\bm{R}^T,
\label{eq:3dgs}
\end{equation}
where the covariance matrix is decomposed as the rotation $\bm{R}$ and scaling $\bm{S}$.
For the rendering process, 3DGS uses a differentiable rasterizer to implement point-based volume rendering. The pixel color $\bm{\mathcal{C}}$ in the image plane is finally rendered by the $\alpha$-blending of the projected Gaussian primitive on this plane.
\begin{equation} \small
    \bm{\mathcal{C}} = \sum^{N}_{i=1}T_i \alpha_i \mathcal{G}_{i}'(\bm{x}; \bm{\mu}_i,\bm{r}_i,s_i) \bm{c}_i, \ \ T_i=\prod_{j \in i-1}(1-\mathcal{G}_j'(\bm{x};\bm{\mu}_i,\bm{r}_i,s_i)\alpha_j) ,
\label{eq:alpha_blending}
\end{equation}
where $T_i$ indicates the accumulated transmittance of $i$-th Guassian at pixel $\bm{x}$. The projected 2D Gaussian probability density $\mathcal{G}'(\bm{x};\bm{\mu},\bm{r},s)$ could be obtained by projection operation from the camera's intrinsic and extrinsic parameters. Finally, the optimizable Gaussian primitives are characterized as: position $\bm{\mu} \in \mathbb{R}^3$, rotation $\bm{r} \in \mathbb{R}^3$, scale factor $s \in \mathbb{R}$, color $\bm{c} \in \mathbb{R}^3$ and opacity $\alpha \in \mathbb{R}$.

\subsection{Gaussian Visibility Control via Learnable Truncation}
\label{sec:truncation}
The clipping plane factor introduces an additional dimension for cinematic rendering baking of 3D medical data with Gaussian splatting, which transfers the 3D representation to the 4D representation problem. Previous works~\cite{diolatzis2024n,liang2024gaufre,wu20244d} try to learn a deformation filed along the introduced dimension to control the rendering of GS. But, these methods conflict with the inspiration that the applied clipping plane mainly changes the visibility of Gaussian primitives during rendering. 
Inspired by the above, we consider the geometric relationship between the clipping plane and Gaussian primitives.

Intuitively, we can directly use the mean value $\bm{\mu}$ as the position of Gaussian to judge the visibility, which we refer to as the Hard Truncation (HT) scheme. Although the rendering results are guaranteed to be generally correct, this simple strategy tends to trap models in local minima, causing artifacts to burr in the clipping surface. In a specific case, we consider the Gaussian primitives, which have an intersection with the clipping plane under $99\%$ confidence interval as per $3\sigma$ principle. They contribute color to the rendering of the clipping surface, but their mean value may fall on both sides of the clipping plane. 
Instead, we consider the real ``barycenter'' of Gaussian actually has an offset $\bm{\delta} \in \mathbb{R}^3$ respects to mean value $\bm{\mu}$. 
\begin{equation} 
    \mathcal{M} = (\bm{\mu} + \bm{\delta}) \cdot \bm{n}  < z ,
\label{eq:vis}
\end{equation}
where $\bm{n}$ is the normal direction of the clipping plane, and $z$ is the distance of the clipping plane to the origin point. $\mathcal{M} \in \{0,1\} $ means the visibility. Gaussians under the clipping plane are visible, and vice versa.

To this end, we propose a Learnable Truncation (LT) scheme by introducing an additional optimizable attribute $m \in \mathbb{R}$ to each Gaussian primitive. 
We formulate  $m = (\bm{\mu} + \bm{\delta}) \cdot \bm{n}$, and rewrite Eq.~\ref{eq:vis} as $ \mathcal{M} = m < z$. 
Thus, we are able to ensure that the operation of $\bm{\mu}$ is not affected by the gradient of Eq.~\ref{eq:vis}.
Inspired by~\cite{lee2024compact}, we employ the straight-through estimator~\cite{bengio2013estimating}, and redefine our learnable truncation function as:
\begin{equation} 
    \mathcal{M} = \operatorname{sg}(\mathds{1}[\sigma(m - z) < \epsilon] - \sigma(m - z)) + \sigma(m - z) ,
\label{eq:truncation}
\end{equation}
where $\sigma(\cdot)$ is the sigmoid function, $\mathds{1}(\cdot)$ is an indicator, $\epsilon$ is a threshold hyperparameter and $\operatorname{sg}(\cdot)$ is the stop gradient operator. In this way, we can directly use the gradient of the sigmoid function to represent the gradient of the visibility step function during backpropagation, and optimizing the value of $m$ automatically.

\subsection{Continuous Clipping via Adaptive Deformation Adjustment}
\label{sec:refinement}
Compared to the 3D scene representation task, interactive cinematic visualization introduces an additional variable to show the internal structures of the volumetric medical data. The introduced dimension transfers this task as a 4D scene representation problem.
But existing GS-based cinematic rendering ~\cite{niedermayr2024application,kleinbeck2024multi} only addresses 3D problems.
We use a novel learnable truncation scheme (Sec.~\ref{sec:truncation}) to control the visibility of Gaussian primitives for cinematic anatomy rendering. However, it is mathematically non-contiguous due to the indicator operation in Eq.~\ref{eq:truncation}, causing inconsistent rendering when the clipping plane changes. Instead of performing a direct visibility control, we propose to predict a continuous adjustment to the Gaussian primitives based on the clipping plane condition, effectively refining the rendering results and ensuring continuity across clipping planes. 

To deal with this issue, we design an adaptive adjustment model to model the dynamic deformation of Gaussians under the specific clipping plane condition. 
We exploit positional encoding (PE)~\cite{mildenhall2021nerf} followed by a tiny MLP to learn a feature $f$, and the desired dynamic deformations are generated from a multi-head MLP decoder. Here, we respectively input all visible Gaussian positions $\bm{\mu}$ and clipping plane parameter $z$ into the PE, and then the merged high-frequency input from PE is fed into the MLP.
\begin{equation} 
    f = \operatorname{MLP}(\gamma(\bm{\mu}), \gamma(z)) ,
\label{eq:adjustment_mlp}
\end{equation}
where $\gamma(\cdot)$ denotes the positional encoding operation.
Then separate MLP heads $\phi_{\mu}$, $\phi_{\gamma}$, and $\phi_{s}$, are employed to compute the deformations of Gaussians: position $\Delta \bm{\mu} = \phi_{\mu} (f)$, rotation $\Delta \bm{r} = \phi_r (f)$ and scale $\Delta \bm{\mu} = \phi_s (f)$.
Finally, we rewrite the rendering function in Eq.~\ref{eq:alpha_blending} as follows:
\begin{equation} 
    \bm{\mathcal{C}} = \sum^{N}_{i=1} \mathcal{M}_i T_i \alpha_i \mathcal{G}_{i}'(\bm{x}; \bm{\mu}_i + \Delta \bm{\mu}_i,\bm{r}_i + \Delta\bm{r}_i, s_i + \Delta s_i) \bm{c}_i .
\label{eq:new_alpha_blending}
\end{equation}
The rendered result is colored by $\alpha$-blending of all visible Gaussians with the query clipping plane.

\subsection{Optimization of ClipGS}
\label{sec:optimization}

We employ a two-step optimization strategy in our method. Firstly, we train the GS with our learnable truncation scheme to obtain a coarse initialization. Then, we start the adaptive adjustment to refine the rendering performance and keep the consistency in the clipping plane dimension.
As for the training loss, we measure the difference between the rendered colors $\bm{C}$ and the ground-truth colors $\bm{C}_\text{gt}$ from the rendered CA image. Similar as 3DGS pipeline in~\cite{kerbl3Dgaussians}, we use the combination of the D-SSIM term and a $\mathcal{L}_1$ term as our training loss:
\begin{equation} \label{eq:loss}
    \mathcal{L} = (1- \lambda) \mathcal{L}_1(\bm{\mathcal{C}}, \bm{\mathcal{C}}_\text{gt}) + \lambda \mathcal{L}_\text{D-SSIM}(\bm{\mathcal{C}}, \bm{\mathcal{C}}_\text{gt}) .
\end{equation}
In this way, we are able to balance the perceptual quality of the rendered images with the fidelity to the real data.
\section{Experiments}
\label{sec:experiments}

\subsection{Experiment Settings}
\boldstartspace{Dataset.}
To evaluate the rendering performance of our proposed method on volume medical data, we generate a custom cinematic medical dataset by BioxelNode~\cite{BioxelNodes} of Blender~\cite{blender}. The custom cinematic medical dataset consists of five scenes: 
1) \textit{Head}; 2) \textit{Lower Limb}; 3) \textit{Skull}; 4) \textit{Thorax}; 5) \textit{Liver} \& \textit{Kidney}.
The original volume of medical data (scenes 1st to 4th) was acquired from the Visible Human Project~\cite{662875}, and the data of scene 5th was acquired from CVH project~\cite{1615382}. We employ a DyNeRF-like~\cite{li2022neural} format in our dataset, which contains a total of 6000 frames with different camera views and timestamps. We rendered 1100 frames for each scene, with random view direction on a unit sphere and random clipping plane position along the z-axis for each frame, all at $800 \times 800$ pixels. We use 1000 frames for training and 100 frames for testing.

\boldstartspace{Implementation Details.}
For each scene, we randomly initialize the GS with 100,000 points within a cube encompassing the scene as default. We train 7000 iterations in the first optimization step, and jointly optimize the GS with our AAM for another 33000 iterations. We employ $\epsilon = 0.5$ in Eq.~\ref{eq:truncation} and $\lambda = 0.2$ in Eq.~\ref{eq:loss}.
For other hyperparameters, we use the same with 3DGS. 
We implement our method based on the PyTorch framework~\cite{paszke2019pytorch} and conduct the experiments with a single RTX 4090 GPU.

\begin{figure}[t]
    \centering
    \includegraphics[width=1\linewidth]{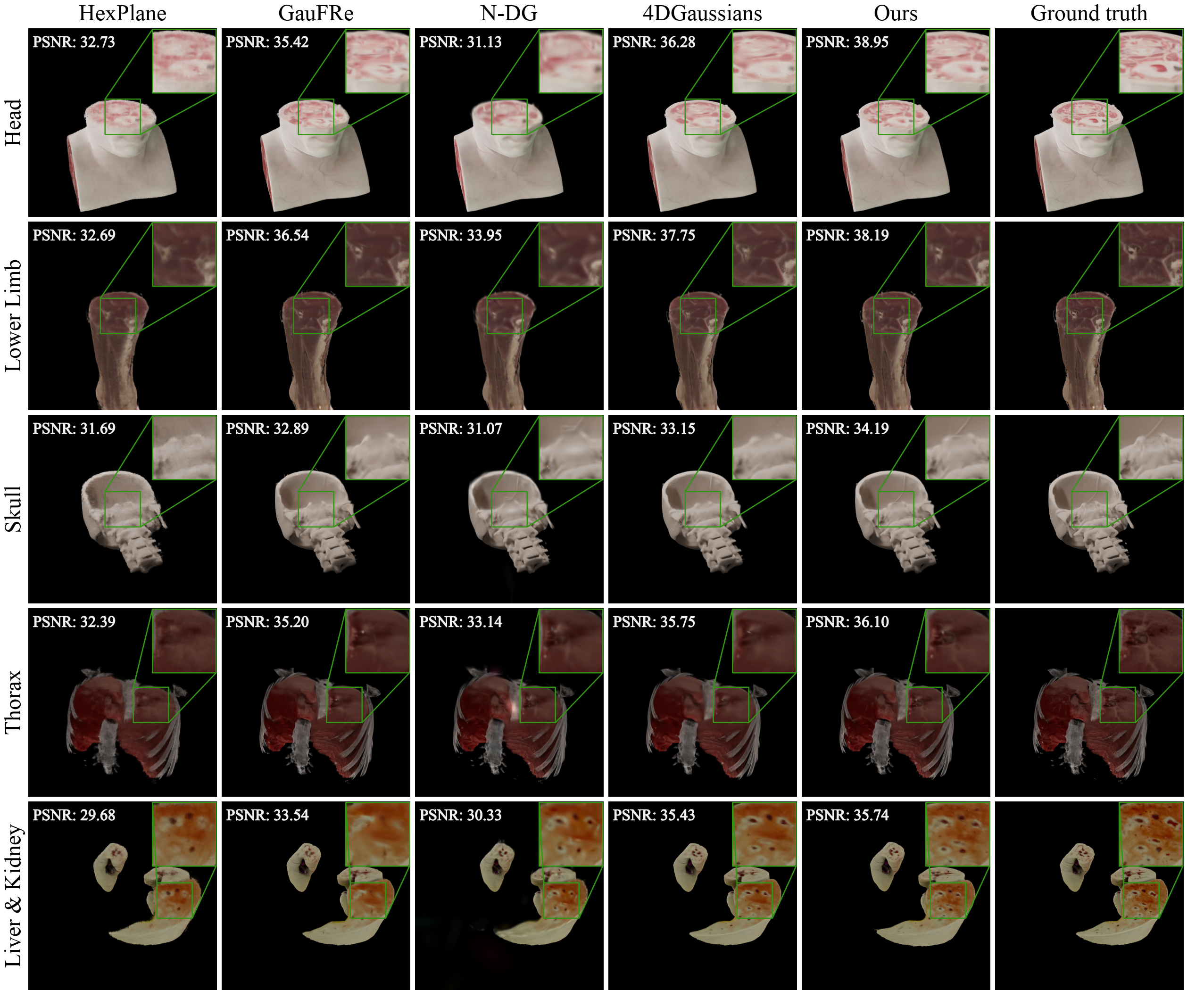}
    \caption{Qualitative results of rendered images on our cinematic medical dataset. The PSNRs in the images denote the rendering quality for the whole case. }
    \label{fig:quantitative}
\end{figure}

\subsection{Qualitative and Quantitative Results}

\boldstartspace{How does our method compare to others?} 
The clipping plane view introduces an additional dimension for the NVS task. Actually, it could be seen as a 4D scene representation problem. Similarly, the clipping plane can be considered as the timestamp in the dynamic scene representation task. 
Thus, we select HexPlane~\cite{cao2023hexplane}, GauFRe~\cite{liang2024gaufre}, N-DG~\cite{diolatzis2024n} and 4DGaussians~\cite{wu20244d} for comparative.
To assess our method's performance, we used Peak Signal-to-Noise Ratio (PSNR), Structural Similarity Index Measure (SSIM) and perceptual quality measure (LPIPS) for image quality, model size for spatial efficiency, and Frames Per Second (FPS) for rendering speed.

\begin{table}[t]
    \caption{\textbf{Quantitative performance.} We compared the results in photometric loss, and also the performance in training time (min), rendering speed, and saving storage on the cinematic anatomy dataset.} 
    \label{tab:quantitative}
    \renewcommand\arraystretch{1.2}
    \centering
    \setlength{\tabcolsep}{1.5mm}
    \scalebox{0.96}{
        \resizebox{\linewidth}{!}{
            \begin{tabular}{@{}c  |c c c |c  |c | c @{}}
                \toprule
                \textbf{Method} &
                \textbf{PSNR}$\uparrow$ & \textbf{SSIM}$\uparrow$ & \textbf{LPIPS}$\downarrow$ &
                \textbf{Time}$ \downarrow$ & \textbf{FPS}$ \uparrow$ &
                \textbf{Storage}$ \downarrow$ \\

                \midrule
                   HexPlane~\cite{cao2023hexplane} &
                31.836 $\pm$ 1.275 & 0.942 $\pm$ 0.015 & 0.109 $\pm$ 0.019 &
                $14.1^\prime$ & 2.7 & 68.0MB\\

                GauFRe~\cite{liang2024gaufre} &
                34.717 $\pm$ 1.481 & \cellcolor{orange}0.968 $\pm$ 0.009 & \cellcolor{orange}0.062 $\pm$ 0.013 &
                $16.6^\prime$ & \cellcolor{orange}150 & \cellcolor{tablered}15.0MB \\

                N-DG~\cite{diolatzis2024n} &
                31.925 $\pm$ 1.541 & 0.937 $\pm$ 0.032 & 0.096 $\pm$ 0.015 &
                $33.6^\prime$ & 129 & 39.7MB\\

                4DGaussians~\cite{wu20244d}&
                \cellcolor{orange}35.653 $\pm$ 1.667 & 0.967 $\pm$ 0.009 & 0.079 $\pm$ 0.014 &
               \cellcolor{tablered}$9.7^\prime$ & 87 & 17.8MB \\
                \textbf{Ours} &
                \cellcolor{tablered}36.635 $\pm$ 1.926 & \cellcolor{tablered}0.974 $\pm$ 0.010 & \cellcolor{tablered}0.061 $\pm$ 0.014 &
                \cellcolor{orange}$13.7^\prime$ & \cellcolor{tablered}156 & \cellcolor{orange}16.1MB \\

                \bottomrule
            \end{tabular}
        }
    }
\end{table}

\boldstartspace{What is our model's performance, and why it works?} 
Our proposed model gives optimal rendering performance among all methods, and also has great performance in training time, reducing storage and also rendering speed, as shown in Tab.~\ref{tab:quantitative} and Fig.~\ref{fig:quantitative}. Other compared methods aim to lean a deformation field to represent the rendering variance along the additional introduced dimension (clipping plane). A significant drawback arises as these methods entail a large model capacity to explain this variance. Instead, we use a novel learnable truncation to incorporate the space geometric relationship between the Gaussian primitives and the clipping plane. Our model allows for a main focus on the visible Gaussian primitive optimization under
a given clipping plane condition, additional AAM to refine the rendering result. 

\subsection{Ablation Study} 

\begin{table}[t] 
    \caption{\textbf{Ablation study of the proposed components.} We compared the effectiveness of our learnable trauncation (LT) and adaptive adjustment model (AAM). Also, we show the superiority of our method compared to the hard truncation (HT) scheme. 3DGS + AAM + LT represents our full model.} 
    \label{tab:ablation}
    \renewcommand\arraystretch{1.2}
    \centering
    \setlength{\tabcolsep}{1mm}
    \scalebox{0.99}{
        \resizebox{\linewidth}{!}{
            \begin{tabular}{@{}c | c c c| c | c | c @{}}
                \toprule
                \textbf{Method} &
                \textbf{PSNR}$\uparrow$ & \textbf{SSIM}$\uparrow$ & \textbf{LPIPS}$\downarrow$ &
                \textbf{Time}$\downarrow$ &
                \textbf{FPS}$\uparrow$ &
                \textbf{Storage}$\downarrow$ \\
                
                \midrule

                3DGS + LT &
                34.577 $\pm$ 1.441& 0.969 $\pm$ 0.012& 0.070 $\pm$ 0.015& \cellcolor{tablered}$11.1^\prime$ & \cellcolor{tablered}201 & \cellcolor{tablered}14.7MB\\

                3DGS + AAM &
                 35.946 $\pm$ 2.043& 0.971 $\pm$ 0.013& 0.064 $\pm$ 0.014& \cellcolor{orange}$11.5^\prime$ & 148 & \cellcolor{orange}15.6MB\\

                3DGS + AAM + HT &
                \cellcolor{orange}36.255 $\pm$ 1.756& \cellcolor{orange}0.971 $\pm$ 0.008& \cellcolor{orange}0.063 $\pm$ 0.014& $14.1^\prime$ & 150 & 15.7MB\\
                
                3DGS + AAM + LT &
                \cellcolor{tablered}36.635 $\pm$ 1.926 & \cellcolor{tablered}0.974 $\pm$ 0.010& \cellcolor{tablered}0.061 $\pm$ 0.014& $13.7^\prime$ & \cellcolor{orange}156 & 16.1MB\\
                
                \bottomrule
            \end{tabular}
        }
    }
\end{table}






\boldstartspace{Learnable Truncation Scheme.} 
The proposed learnable truncation allows our model to focus mainly on the visible Gaussian primitive optimization under a given clipping plane condition. Compared to the quantitative results (3DGS + AAM vs. 3DGS + AAM + LT) in Tab.~\ref{tab:ablation}, our LT scheme shows efficient performance in reducing the floaters, training time, saving storage, and improving details.
Besides, we also compared our Learnable truncation with the Hard truncation, which has been mentioned in Sec.~\ref{sec:truncation} (3DGS + AAM + LT vs. 3DGS + AAM + HT); HT has almost the same performance in reducing training time and saving the storage. However, the non-optimizable strategy compromises the rendering quality of the model.

\boldstartspace{Adaptive Adjustment Model.} 
Our adaptive adjustment model predicts a dynamic deformation for the Gaussian. So that the position, rotation and scaling of these Gaussian primitives could be adaptively adjusted given the specific clipping plane. 
By incorporating an additional MLP to learn the clipping plane-dependent variations, our proposed module exhibits greater capacity to mitigate photometric loss, thereby achieving improved results, as shown in Table~\ref{tab:ablation} (3DGS + LT vs. 3DGS + AAM + LT).

\section{Conclusion}

In this paper, we proposed a novel GS-based framework
that can render volumetric medical data with clipping planes with photorealism and real-time. 
We propose a learnable truncation scheme that automatically adjusts the visibility of Gaussian primitives in response to the clipping plane. 
In addition, we propose an adaptive adjustment model to model the dynamic deformation of Gaussians under the specific clipping plane condition. The experiments on our cinematic medical dataset demonstrate that our method achieves superior image quality, rendering speed, and storage efficiency compared to existing approaches, making it a promising solution for more accessible and interactive medical imaging applications in surgical planning and medical education.





\bibliographystyle{splncs04}
\bibliography{refs}

\end{document}